\documentclass[conference]{IEEEtran}
\IEEEoverridecommandlockouts

\usepackage{cite}
\usepackage[T1]{fontenc}
\usepackage{textcomp}
\usepackage{amsmath,amssymb,amsfonts}
\usepackage{mathrsfs}
\usepackage{latexsym}
\usepackage{algorithm}
\usepackage{algorithmic}
\usepackage{xspace}
\usepackage{graphicx}
\usepackage{epsfig}
\usepackage{float}
\usepackage{array}
\usepackage{booktabs}
\usepackage{multirow}
\usepackage{threeparttable}
\usepackage{caption}
\usepackage{subcaption}
\usepackage{xcolor}
\usepackage{url}
\usepackage[colorlinks,
pagebackref=false,breaklinks=true,
            linkcolor=red,
            anchorcolor=red,
            urlcolor=red,
            hyperfootnotes=true,
            citecolor=green]{hyperref}

\newcommand{\bheading}[1]{{\noindent{\textbf{#1}}}}

\newcommand{\tabincell}[2]{\begin{tabular}{@{}#1@{}}#2\end{tabular}}

\def\BibTeX{{\rm B\kern-.05em{\sc i\kern-.025em b}\kern-.08em
    T\kern-.1667em\lower.7ex\hbox{E}\kern-.125emX}}
\newcommand{\algname}{\textsc{SGB-Match}\xspace}

\begin{document}

\title{Structure-Guided Self-Supervised Matching for One-Shot Medical Landmark Detection}


\author{\IEEEauthorblockN{1\textsuperscript{st} Qingsong Yao}
\IEEEauthorblockA{\textit{Soochow School of Artificial Intelligence} \\
\textit{Renmin University of China}\\
qingsongyao98@gmail.com}
\and
\IEEEauthorblockN{2\textsuperscript{nd} Zhen Huang}
\IEEEauthorblockA{\textit{Department of Computer Science)} \\
\textit{University of Science and Technology of China}\\
}
\and
\IEEEauthorblockN{3\textsuperscript{rd} Ao Wang}
\IEEEauthorblockA{\textit{Soochow School of Artificial Intelligence}\\
\textit{Renmin University of China}\\
}
\and
\IEEEauthorblockN{4\textsuperscript{th} Rongsheng Wang}
\IEEEauthorblockA{\textit{School of Biomedical Engineering)} \\
\textit{University of Science and Technology of China}}
\and
\IEEEauthorblockN{5\textsuperscript{th} Hanxue Zhang}
\IEEEauthorblockA{\textit{School of Artificial Intelligence} \\
\textit{Shanghai Jiaotong University}}
\and
\IEEEauthorblockN{6\textsuperscript{th} Jianji Wang}
\IEEEauthorblockA{\textit{Department of Orthopedics} \\
\textit{Affiliated Hospital of Guizhou Medical University}\\
\textit{Yale University}\\
Jianji.wang@yale.edu}
\and
\IEEEauthorblockN{7\textsuperscript{th} S. Kevin Zhou}
\IEEEauthorblockA{\textit{School of Biomedical Engineering} \\
\textit{University of Science and Technology of China}\\
s.kevin.zhou@gmail.com}
}

\maketitle

\begin{abstract}
Medical landmark detection usually requires accurate expert annotations, which are laborious and difficult to scale across anatomical regions.
In this work, we study an extreme annotation-efficient setting where only a single annotated template image is available.
We propose \algname, a structure-guided coarse-to-fine self-supervised matching framework for one-shot medical landmark detection.
The framework first learns dense anatomical correspondence from unlabeled augmented image pairs, and then transfers the landmark definition from the annotated template to each target image through feature matching.
Different from standard contrastive correspondence learning, where negative candidates are penalized by a structure-agnostic rule, we introduce a structure-guided bias into the contrastive objective.
The bias is constructed from relative distance and edge-aware anatomical cues, and explicitly reweights the negative gradients: nearby structure-relevant candidates are weakly repelled, while distant or structure-irrelevant negatives are strongly suppressed.
As a result, the learned feature space better preserves local anatomical structures around template landmarks and reduces confusing responses from repeated textures.
We further adopt a global-to-local design, where a global encoder provides coarse landmark localization and a local encoder refines the prediction in a cropped region.
Extensive experiments on four 2D radiological landmark datasets demonstrate that \algname achieves strong one-shot performance across both public and newly collected datasets and consistently benefits from both structure-guided bias and two-stage refinement.

\end{abstract}


\section{Introduction}

Anatomical landmark detection is a fundamental problem in medical image analysis. 
Clinically defined landmarks provide compact geometric descriptions of anatomical structures and support a wide range of downstream applications, including cephalometric analysis, skeletal maturity assessment, chest structure measurement, orthopedic alignment evaluation, and surgical planning.
Early landmark localization methods relied on handcrafted appearance descriptors, statistical shape models, random forests, and constrained geometric priors~\cite{cootes1995active,lindner2014robust,ref_urschler,huang2025h3de}.
With the development of deep learning, fully supervised landmark detectors have achieved strong performance by learning heatmap regression, coordinate refinement, spatial configuration modeling, graph-based anatomical reasoning, and uncertainty-aware representations from large annotated datasets~\cite{ref_scn,chen2019cephalometric,Wei2020Measurements,li2020structured,mccouat2022contour,zhu2021you, zhou2024hybrid}.
These methods demonstrate the potential of deep models for accurate and robust medical landmark localization.

However, the success of supervised landmark detection is tightly coupled with dense expert annotation.
Unlike image-level labels, landmark annotations require experts to place each point at a precise anatomical position, which is time-consuming, observer-dependent, and difficult to scale across anatomical regions and imaging protocols.
Semi-supervised learning can reduce annotation cost by exploiting unlabeled images, but most semi-supervised methods still require a reliable initial detector trained from a non-trivial amount of labeled data~\cite{yang2021survey,sohn2020fixmatch,chen2022semi, he2025landmarks}.
This requirement becomes problematic in new clinical applications, where only a few annotated images, or even a single annotated template, may be available.
Therefore, one-shot medical landmark detection has attracted increasing attention as an extreme annotation-efficient setting, where the landmark semantics are specified by only one annotated image~\cite{quan2021images,yao2021one,lei2021contrastive,yan2022sam,yin2022one,zhu2023uod}.

A common strategy for one-shot landmark detection is to learn anatomical correspondence from unlabeled images and transfer landmark definitions from the annotated template to target images.
For example, Cascade Comparing to Detect (CC2D) learns cascade correspondence through self-supervised matching between augmented image pairs~\cite{yao2021one}; RPR-Net learns relative position regression for one-shot localization~\cite{lei2021contrastive}; SAM learns pixel-wise anatomical embeddings in radiological images~\cite{yan2022sam};  and UOD further studies universal one-shot landmark representations across anatomical regions~\cite{zhu2023uod}.
Despite their progress, existing correspondence-based methods are still limited by the structure-agnostic nature of their contrastive objectives.
In standard point-level contrastive learning, the positive correspondence is pulled closer in feature space, while all negative candidates are penalized under the same generic softmax formulation.
Although their gradient magnitudes depend on feature similarity, the loss itself does not distinguish whether a negative candidate is anatomically plausible or not.
Consequently, a point on the same local anatomical structure and a point on an irrelevant repeated texture may be treated by the same structure-agnostic rule.

We argue that one-shot medical landmark detection should not be treated as generic pixel matching.
Medical landmarks are usually structure-anchored points: they are defined by local anatomical geometry, such as boundaries, contour intersections, skeletal axes, or extremal points of clinically meaningful structures.
Therefore, the feature around a template landmark should be encouraged to stay close not only to its exact positive correspondence, but also to nearby structure-relevant anatomical textures, while being separated from distant or off-structure candidates.
This observation motivates a structure-guided contrastive formulation.
Instead of penalizing all negatives in a structure-agnostic manner, we introduce candidate-specific bias terms that reshape the effective contrastive gradients according to anatomical distance and edge-aware structural relevance.

To address this issue, we first introduce a \textbf{structure-guided bias} for contrastive self-supervised learning.
The bias is constructed from two complementary cues: a distance-aware term and an edge-aware term.
The distance-aware term assigns larger bias to candidates farther away from the positive point, while the edge-aware term penalizes candidates that are not located on anatomical boundaries.
By adding this candidate-specific bias to the contrastive logit, the loss no longer treats negative candidates only through a structure-agnostic softmax form.
Instead, nearby structure-relevant candidates receive weaker repulsion, while distant or off-structure negatives receive stronger gradients and are pushed away more aggressively.
We further provide a gradient-level derivation showing that the negative gradient is scaled by an exponential factor induced by the bias term.
This explains why the proposed objective encourages the encoder to learn structure-aware anatomical embeddings rather than arbitrary texture responses.

Based on the structure-guided bias, we further design a \textbf{coarse-to-fine self-supervised matching} strategy.
The global encoder is trained on resized full images to learn long-range anatomical correspondence and produce coarse landmark localization.
The local encoder is then trained on landmark-centered crops to refine the prediction using local structural evidence.
Importantly, the coarse predictions are used only as crop anchors and sampling priors for the fine stage, rather than as ground-truth pseudo heatmaps.
During inference, a single annotated template landmark is first matched to the target image by the global encoder to obtain a coarse location, and the local encoder then performs fine-grained matching within the cropped region.

Putting these components together, we propose \algname, a \textbf{structure-guided global-to-local matching} framework for one-shot medical landmark detection.
The proposed method learns dense anatomical correspondence from unlabeled images and transfers landmark definitions from a single annotated template to each target image.
Different from previous self-supervised matching methods such as CC2D~\cite{yao2021one}, \algname explicitly injects anatomical structure into the contrastive objective and uses a dedicated two-stage inference process to progressively refine landmark localization.

We evaluate \algname on four 2D radiological landmark datasets, including Cephalometric X-ray~\cite{wang2016benchmark}, Hand X-ray~\cite{ref_scn}, Chest X-ray~\cite{zhu2021you}, and a newly collected lower-extremity X-ray dataset for biomechanical parameter measurement.
The new BMPLE dataset contains lower-limb radiographs with clinically defined landmarks for orthopedic alignment evaluation, which is important for measuring varus/valgus deformity and planning osteotomy-related procedures~\cite{teeter2018varus}.
Across these datasets, \algname achieves state-of-the-art one-shot performance, and ablation studies verify the effectiveness of both the structure-guided bias and the local refinement stage.

Our contributions are summarized as follows:
\begin{itemize}
    \item We formulate one-shot medical landmark detection as \textbf{structure-guided anatomical correspondence learning} and propose \algname, a structure-guided global-to-local matching framework validated on four 2D radiological landmark benchmarks.
    
    \item We introduce a \textbf{structure-guided bias} for self-supervised dense matching. Unlike standard contrastive objectives that optimize negative candidates in a structure-agnostic manner, the proposed bias reweights negative gradients according to distance and edge-aware anatomical relevance, preserving nearby structure-relevant candidates while suppressing implausible negatives.
    
    \item We contribute BMPLE, a lower-extremity X-ray landmark dataset for biomechanical parameter measurement. 
\end{itemize}

\begin{figure*}[htbp]
\centering
\includegraphics[width=\textwidth]{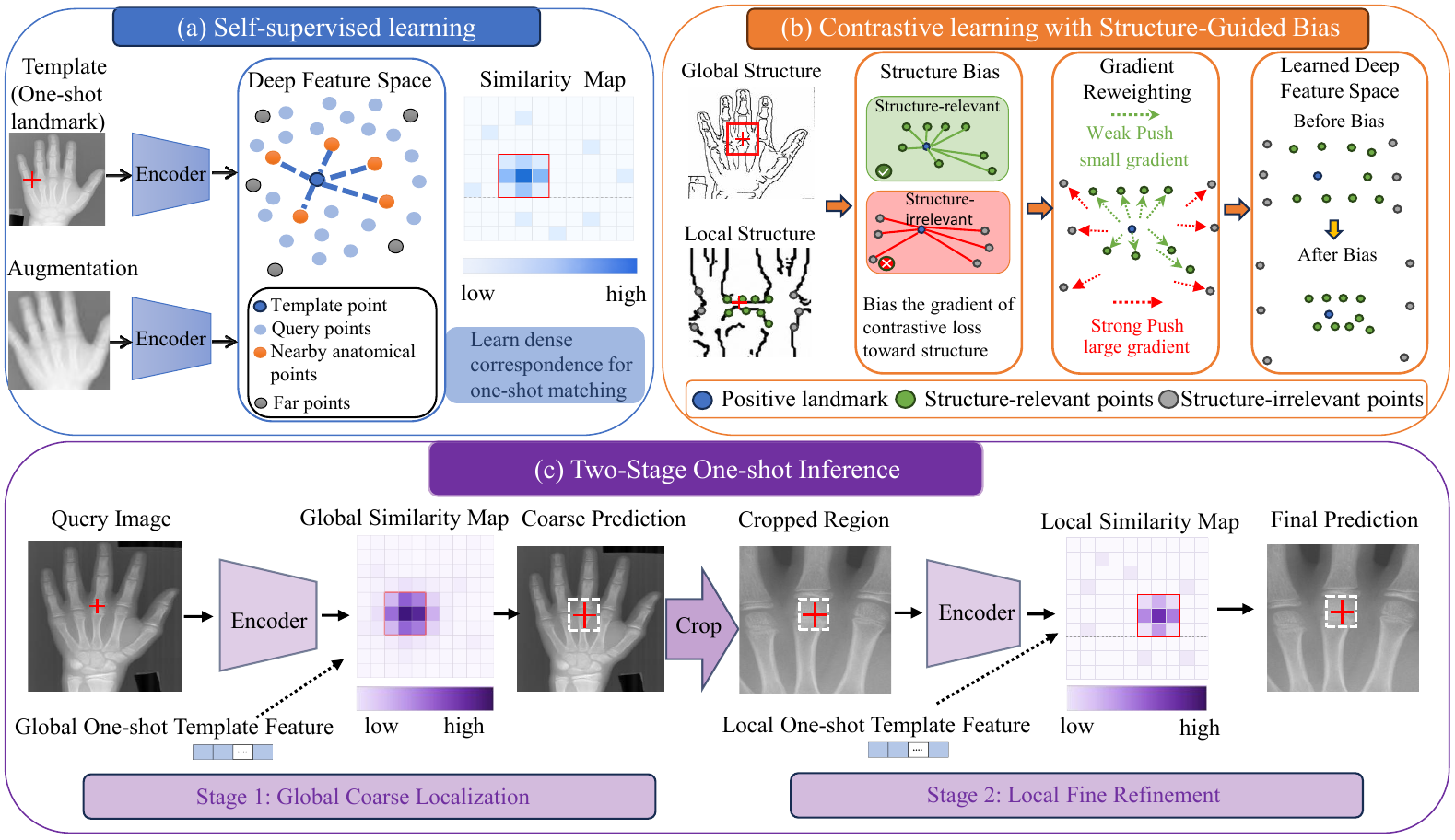}
\caption{
Overview of the proposed structure-guided coarse-to-fine one-shot landmark detection framework (\algname).
(a) Dense anatomical correspondence is learned by self-supervised contrastive learning from augmented image pairs, enabling a single annotated template landmark to retrieve its corresponding location through feature matching.
(b) The structure-guided bias reweights contrastive gradients by anatomical relevance, weakly repelling nearby structure-relevant candidates while strongly suppressing structure-irrelevant negatives, thereby shaping a structure-aware feature space.
(c) The framework is trained with separate global and local encoders. During inference, the global encoder first produces a coarse prediction, and the local encoder refines it within the cropped region for final landmark localization.
}
\label{fig:main}
\end{figure*}

\section{Related Work}

\subsection{Self-supervised and contrastive learning}

Self-supervised learning has been widely used to learn visual representations from unlabeled data.
Representative contrastive methods, such as MoCo, SimCLR, BYOL, and Barlow Twins, learn invariant image-level representations by comparing different augmented views~\cite{he2020momentum,chen2020simple,grill2020bootstrap,zbontar2021barlow}.
In medical image analysis, self-supervised learning has also been explored through image restoration, anatomical transformation prediction, superpixel-level learning, and patch-wise contrastive learning~\cite{zhou2019models,zhou2021models,zhu2020rubik,ouyang2020self,chaitanya2020contrastive}.
These methods show that unlabeled medical images contain useful anatomical priors.

However, landmark detection requires spatially accurate point-level representations rather than only image-level semantics.
Recent works therefore study dense or pixel-wise self-supervised learning for anatomical correspondence~\cite{yan2022sam,yao2021one}.
A recent diffusion-based pre-training method also shows that task-specific self-supervision can benefit few-shot X-ray landmark detection~\cite{di_via_2025_wacv}.
Our method is inspired by dense contrastive correspondence learning, but further introduces a structure-guided bias so that the matching objective explicitly considers anatomical boundaries and local structural cues.

\subsection{Fully supervised medical landmark detection}

Fully supervised landmark detection has achieved strong performance with sufficient expert annotations.
Early methods used shape models, random forests, and regression-voting strategies to localize anatomical points~\cite{cootes1995active,lindner2014robust,ref_urschler}.
Deep learning methods further improve accuracy by predicting heatmaps, coordinate offsets, or regression targets from convolutional features~\cite{payer2016regressing,ref_scn,zhong2019attention,chen2019cephalometric}.
To model anatomical relationships, later methods incorporate spatial configuration, anatomical constraints, graph reasoning, contour-aware heatmaps, and universal multi-dataset training~\cite{Wei2020Measurements,li2020structured,mccouat2022contour,zhu2021you}.

Although effective, these methods usually require many annotated images for each target anatomy.
This requirement limits their use in new clinical scenarios, where landmark definitions may be task-specific and expert annotation is expensive.
In contrast, this work focuses on the one-shot setting, where only one annotated template image is available.

\subsection{One-shot medical landmark detection}

One-shot medical landmark detection aims to localize landmarks using only a single annotated template.
A common solution is to learn anatomical correspondence from unlabeled images and transfer the landmark definitions from the template to target images.
CC2D learns cascade self-supervised correspondence between augmented image pairs~\cite{yao2021one}.
RPR-Net learns relative position regression for one-shot localization~\cite{lei2021contrastive}.
SAM learns pixel-wise anatomical embeddings in radiological images~\cite{yan2022sam}.
Yin \textit{et al.} introduce edge-guided transform and noisy landmark refinement~\cite{yin2022one}.
UOD further studies universal one-shot landmark detection across different anatomical regions~\cite{zhu2023uod}.

Recent methods also introduce foundation models into annotation-efficient landmark detection.
FM-OSD uses a frozen visual foundation model with global and local feature decoders for one-shot landmark detection~\cite{miao2024fmosd}.
GeoSapiens adapts a human-centric foundation model for few-shot dental landmark detection with a geometric loss~\cite{wang2025geosapiens}.
KAN-OSD combines a DINO-based encoder with KAN-based decoders for one-shot anatomical landmark detection~\cite{tian2026kanosd}. Different from these methods, our work focuses on structure-guided self-supervised anatomical matching.
Instead of relying only on appearance similarity, distance-aware matching, or pseudo-label refinement, we explicitly inject anatomical structure into the contrastive objective.

\section{Method}

\subsection{CC2D Contrastive Learning Framework}
\label{Sec:cc2d_framework}

We study one-shot medical landmark detection with one annotated template image and a set of unlabeled target images.
Let the annotated template be $T$ with $K$ landmarks $P^T=\{p_k^T\}_{k=1}^{K}$.
Given a target image $X$, the goal is to estimate the corresponding landmarks $P^X=\{p_k^X\}_{k=1}^{K}$.
Our method, \algname, builds on the CC2D contrastive learning framework~\cite{yao2021one}, and introduces two new components: a structure-guided bias for contrastive learning and a global-to-local refinement stage.

We first briefly define the CC2D framework~\cite{yao2021one}.
Given an unlabeled image $X$, an augmented view $X'$ is generated by known spatial transformations.
For a sampled point $p$ in $X$, its corresponding point $p'$ in $X'$ is known from the transformation.
A dense feature extractor $F_{\theta}=\{f_{\theta}^{i}\}_{i=1}^{L}$ produces multi-scale feature maps.
At feature level $i$, the coordinates of $p$ and $p'$ are denoted as $p_i$ and $p_i'$.
CC2D learns point-level contrastive representations by comparing the feature of $p_i'$ in $X'$ with candidate positions in $X$:
\begin{equation}
s^i(q,p_i')
=
\cos\!\left(
f_{\theta}^{i}(X)(q),
f_{\theta}^{i}(X')(p_i')
\right).
\label{Eq:cos}
\end{equation}

CC2D performs contrastive learning across multiple scales:
\begin{itemize}
    \item deep feature maps provide coarse anatomical context;
    \item shallow feature maps preserve local structural details;
    \item template-to-target inference multiplies similarity maps from different scales.
\end{itemize}
For readability, we write ${\cal M}_i={\cal M}_i(p_i)$.
Without additional structure guidance, the candidate probability and base contrastive loss are
\begin{equation}
\begin{aligned}
\hat{h}^i(q)
&=
\frac{\exp(\tau s^i(q,p_i'))}
{\sum_{\bar q\in{\cal M}_i}
\exp(\tau s^i(\bar q,p_i'))},\\
\hat{\mathcal{L}}^i
&=
-\log \hat{h}^i(p_i),
\end{aligned}
\label{Eq:base_loss}
\end{equation}
where $\tau$ is the temperature and the positive target is the same physical point $p_i$.

During one-shot inference, the annotated template is used as the query source.
For landmark $k$, the template feature at $p_k^T$ is compared with candidate positions in the target image $X$ at each feature level.
The similarity maps are upsampled to the image grid and multiplied, and the landmark prediction is obtained by
\begin{equation}
\begin{aligned}
S_k(p|T,X)
&=
\prod_i
\operatorname{clip}_{[0,1]}
\!\left\{
s^i[p_k^T,p|T,X]
\right\},\\
\hat{p}_k
&=
\arg\max_p S_k(p|T,X).
\end{aligned}
\label{Eq:cascade_inference}
\end{equation}

Although CC2D learns dense anatomical representations from unlabeled images, its contrastive objective mainly depends on feature similarity and spatial search range.
It does not explicitly consider whether a negative candidate lies on an anatomically meaningful structure.
Therefore, we introduce a structure-guided bias into the contrastive objective.

\begin{figure}
    \centering
    \includegraphics[width=1\linewidth]{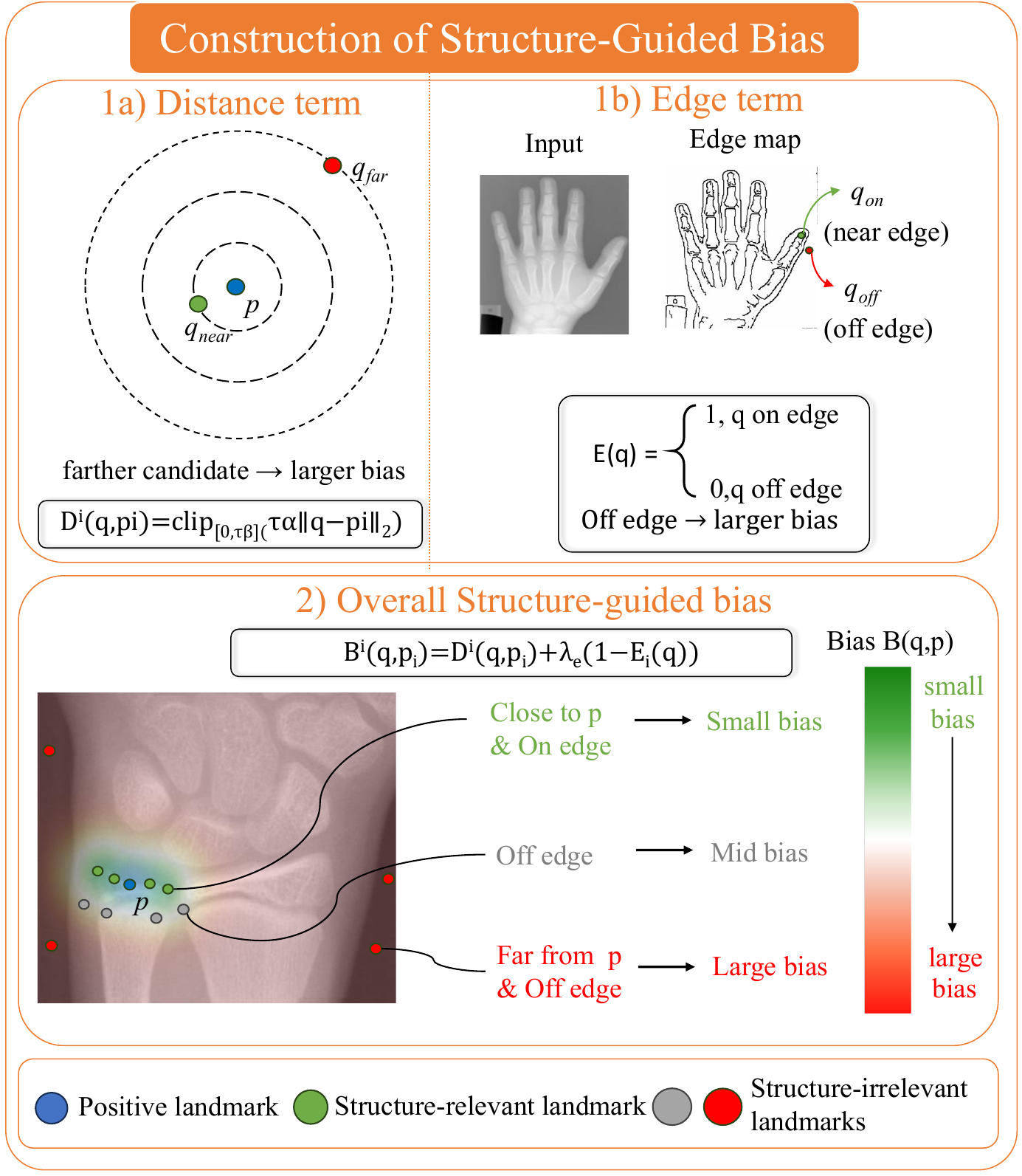}
    \caption{Distance term and edge term in the construction of structure-guided bias, which encourages higher similarity for more structure-relevant landmark pairs.}
    \label{fig:SGB}
\end{figure}

\subsection{Structure-Guided Bias}
\label{Sec:sgb}

Medical landmarks are usually anchored by anatomical geometry, such as cortical boundaries, rib contours, joint edges, skeletal axes, or contour intersections.
A misleading negative candidate should be suppressed not only when it is far from the positive point, but also when it is inconsistent with local anatomical structures.
We encode this intuition by a \textbf{structure-guided bias} (SGB).

For a candidate $q$, we first define a distance term:
\begin{equation}
D^i(q,p_i)
=
\operatorname{clip}_{[0,\tau\beta]}
\!\left(
\tau\alpha\|q-p_i\|_2
\right),
\label{Eq:distance_bias}
\end{equation}
where $\alpha$ controls the slope and $\beta$ controls the maximum bias.
Then, an edge map $E^i(q)\in\{0,1\}$ is used to indicate whether $q$ lies on local anatomical boundaries.
To avoid an overlong expression, we define the edge penalty as
\begin{equation}
P_e^i(q)
=
\mathbf{1}_{i\in{\cal I}_{e}}
\tau\lambda_e(1-E^i(q)),
\label{Eq:edge_penalty}
\end{equation}
where $\lambda_e$ controls the edge penalty and ${\cal I}_{e}$ denotes the feature levels where edge guidance is used.
The complete SGB is defined as
\begin{equation}
B^i(q,p_i)
=
\begin{cases}
0, & q=p_i,\\[1mm]
\operatorname{clip}_{[0,\tau\beta]}
\!\left(D^i(q,p_i)+P_e^i(q)\right),
& q\neq p_i.
\end{cases}
\label{Eq:sgb}
\end{equation}
The positive point has zero bias.
Farther negatives and non-edge negatives receive larger bias.

We add SGB to the contrastive logit:
\begin{equation}
w^i(q)
=
\tau s^i(q,p_i')
+
B^i(q,p_i).
\label{Eq:biased_logit}
\end{equation}
The biased probability and structure-guided contrastive objective are
\begin{equation}
\begin{aligned}
h^i(q)
&=
\frac{\exp(w^i(q))}
{\sum_{\bar q\in{\cal M}_i}\exp(w^i(\bar q))},\\
\mathcal{L}_{SSL}
&=
\sum_i -\log h^i(p_i).
\end{aligned}
\label{Eq:ssl_loss}
\end{equation}

\noindent\textbf{Gradient analysis.}
We now analyze how SGB changes the optimization of positive and negative candidates.
For the positive point $p_i$, because $B^i(p_i,p_i)=0$, we have
$\partial \mathcal{L}_{SSL}^{i}/\partial s^i(p_i,p_i')<0$.
Thus, gradient descent increases the similarity of the positive pair.

For a negative candidate $\hat{q}\neq p_i$, let
$Z^i=\sum_{q\in{\cal M}_i}\exp(w^i(q))$.
The gradient is
\begin{equation}
\begin{aligned}
\frac{\partial \mathcal{L}_{SSL}^{i}}
{\partial s^i(\hat q,p_i')}
&=
\tau h^i(\hat q)\\
&=
\frac{\tau}{Z^i}
\exp\!\left(\tau s^i(\hat q,p_i')\right)
\exp\!\left(B^i(\hat q,p_i)\right).
\end{aligned}
\label{Eq:negative_gradient}
\end{equation}
This gradient is positive, so gradient descent decreases the similarity of the negative pair.
Moreover, the magnitude is proportional to $\exp(B^i(\hat q,p_i))$.
Thus, a negative candidate with larger SGB is pushed downward more strongly.

For two negative candidates $q_a$ and $q_b$, their gradient ratio is
\begin{equation}
\begin{aligned}
&
\frac{
\partial \mathcal{L}_{SSL}^{i}/\partial s^i(q_a,p_i')
}{
\partial \mathcal{L}_{SSL}^{i}/\partial s^i(q_b,p_i')
}
\\
&=
\exp\!\left(
\tau[s^i(q_a,p_i')-s^i(q_b,p_i')]
\right)
\exp\!\left(
B^i(q_a,p_i)-B^i(q_b,p_i)
\right).
\end{aligned}
\label{Eq:gradient_ratio}
\end{equation}
When two negatives have similar feature similarity, the ratio is mainly determined by the bias difference:
\begin{equation}
\frac{
\partial \mathcal{L}_{SSL}^{i}/\partial s^i(q_a)
}{
\partial \mathcal{L}_{SSL}^{i}/\partial s^i(q_b)
}
\approx
\exp\!\left(
B^i(q_a,p_i)-B^i(q_b,p_i)
\right).
\label{Eq:gradient_ratio_simple}
\end{equation}
Therefore, farther negatives and non-edge negatives receive exponentially stronger suppression, while nearby edge-aligned candidates are not over-penalized.
This explains why SGB guides the contrastive learning process toward anatomical structures rather than arbitrary texture responses.

\subsection{Global-to-Local Refinement}
\label{Sec:global_local}

The CC2D framework performs template-to-target localization on the full image.
It provides a coarse landmark location, but local ambiguity may remain when repeated edges or similar anatomical structures appear.
We therefore introduce a global-to-local refinement stage.

In the first stage, the global contrastive learning framework predicts a coarse landmark center:
\begin{equation}
\hat{p}_{k}^{coarse}
=
\arg\max_p S_k(p|T,X).
\label{Eq:coarse_prediction}
\end{equation}
This prediction is used only to define a local crop and search prior.
It is not treated as a ground-truth pseudo heatmap.

In the second stage, for landmark $k$, we crop a target patch $X_k^{crop}$ centered at $\hat{p}_{k}^{coarse}$ and a template patch $T_k^{crop}$ centered at $p_k^T$.
A local contrastive learning framework is trained with the same SGB-based objective in Eq.~\ref{Eq:ssl_loss}.
At inference, the local score is
\begin{equation}
S_k^{crop}(p)
=
\prod_i
\operatorname{clip}_{[0,1]}
\!\left\{
s^i[p_k^{T,crop},p|T_k^{crop},X_k^{crop}]
\right\}.
\label{Eq:local_score}
\end{equation}
The crop-level prediction is
\begin{equation}
\hat{p}_k^{crop}
=
\arg\max_{p\in{\cal R}_k}
S_k^{crop}(p),
\label{Eq:local_prediction}
\end{equation}
where ${\cal R}_k$ is the local search region.
Finally, the crop prediction is mapped back to the original image:
\begin{equation}
\hat{p}_k^{X}
=
r_k\hat{p}_k^{crop}+o_k,
\label{Eq:back_projection}
\end{equation}
where $o_k$ is the crop origin and $r_k$ is the resize ratio.
The global stage handles large anatomical variation, while the local stage refines the prediction using nearby structural evidence.

\begin{figure}[h]\centering\includegraphics[width=0.5\textwidth]{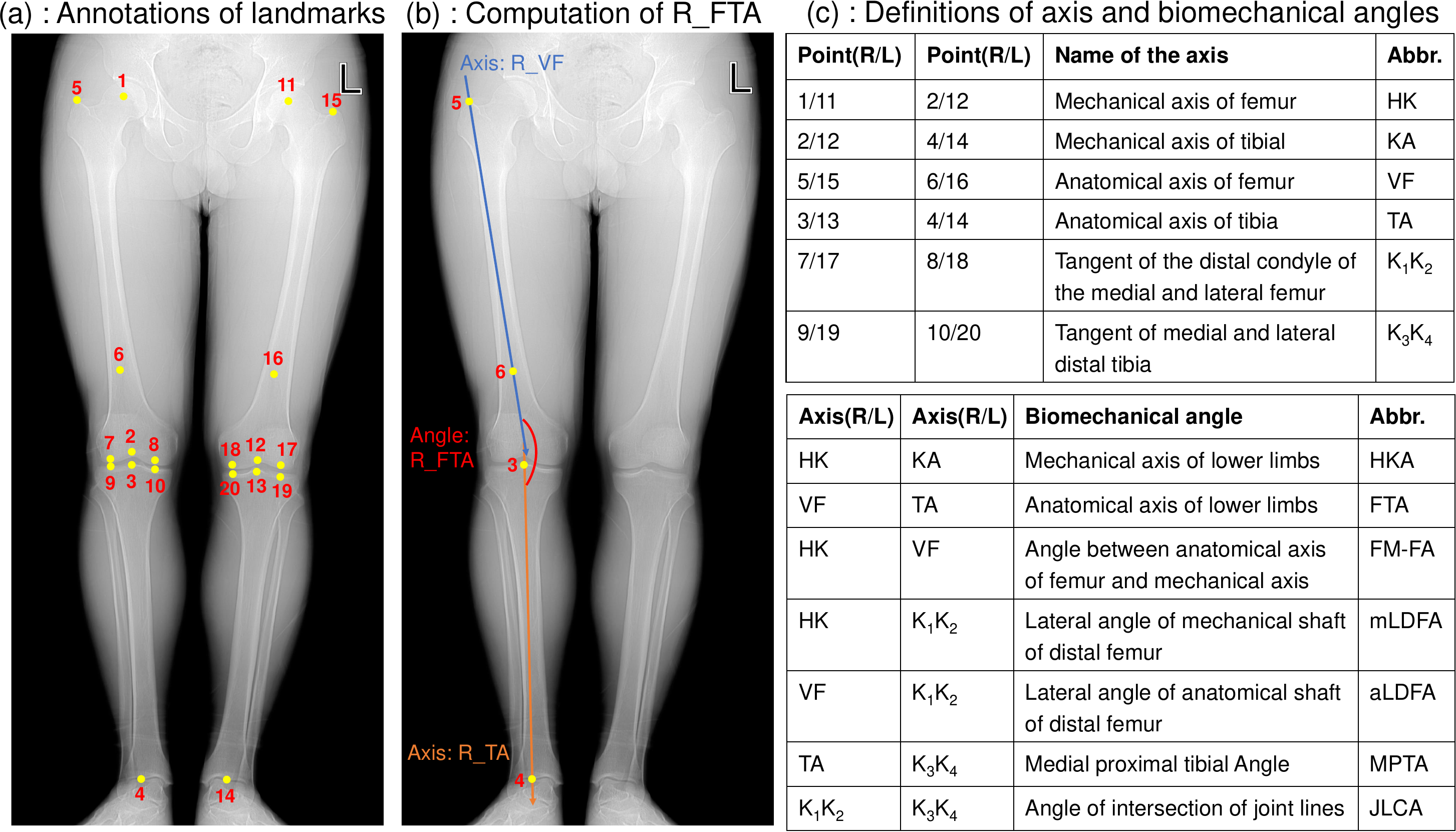}\caption{(a) The illustration of the 20 annotated landmarks in the lower extremity radiography in BMPLE dataset, 1-10 landmarks are on the right (R) lower limb, and landmarks 11–20 are on the left (L) lower limb. (b) The biomechanical angle (Right FTA in figure b) is calculated between two axes, each axis is connected by two landmarks. (c) The detailed definitions of axis and biomechanical angles. } 
\label{fig:dataset}\end{figure}

\begin{table*}[h]
\centering
\footnotesize
\caption{Comparison of supervised approaches and annotation-efficient methods on Head, Hand, Chest, and Leg test sets. * denotes results copied from original papers, and $\#$ denotes our re-implemented results. DDPM reports the 1-shot supervised fine-tuning result from~\cite{di_via_2025_wacv}; it is included as an annotation-efficient reference but is not a template-based one-shot matching method. Ours-Coarse w/o bias removes both distance and edge-aware bias, Ours-Coarse w/o edges keeps the distance bias only, Ours-Coarse uses the full structure-guided bias, and Ours-Fine denotes the final global-to-local refinement result. The best one-shot/template-based result in each dataset block is in \textbf{bold}.}
\begin{threeparttable}
\begin{tabular}{l|l|l|c|ccccc}
\bottomrule
\multirow{2}{*}{Dataset} & \multirow{2}{*}{Model} & \multirow{2}{*}{Conf./Jour.} & \multirow{2}{*}{\tabincell{c}{Labeled\\images}} & \multirow{2}{*}{\tabincell{c}{MRE ($\downarrow$)\\(mm/px)}} & \multicolumn{4}{c}{SDR ($\uparrow$) (\%)} \\
\cline{6-9}
 & & & & & 2 mm & 2.5 mm & 3 mm & 4 mm \\
\hline

\multirow{13}{*}{Head}
& McCouat \textit{et al.}~\cite{mccouat2022contour}* & CVPR-22 & 150 & 1.30 & 79.93 & 86.92 & 90.44 & 95.41 \\
\cline{2-9}
& RPR-Net~\cite{lei2021contrastive}\# & MICCAI-21 & 1 & 4.45 & 19.45 & 26.40 & 36.06 & 52.74 \\
& SAM~\cite{yan2022sam}\# & TMI-22 & 1 & 2.56 & 54.11 & 63.66 & 70.25 & 80.84 \\
& Yin \textit{et al.}~\cite{yin2022one}* Stage I & ECCV-22 & 1 & 2.70 & 42.78 & 54.88 & 65.03 & 81.01 \\
& Yin \textit{et al.}~\cite{yin2022one}* Stage II & ECCV-22 & 1 & 2.13 & 54.69 & 67.47 & 77.85 & 90.02 \\
& UOD~\cite{zhu2023uod}* & MICCAI-23 & 1 & 2.43 & 51.14 & 62.37 & 74.40 & 86.49 \\
& DDPM~\cite{di_via_2025_wacv}* & WACV-25 & 1 & 15.71 & 17.31 & 27.14 & 33.24 & 45.14 \\
& FM-OSD~\cite{miao2024fmosd}* & MICCAI-24 & 1 & 1.82 & 67.35 & 77.92 & \textbf{84.59} & \textbf{91.92} \\
& KAN-OSD~\cite{tian2026kanosd}* & ISBI-26 & 1 & 2.06 & 60.65 & 72.76 & 81.44& 90.12 \\
& Ours-Coarse w/o bias & ours & 1 & 2.09 & 64.21 & 71.52 & 79.20 & 87.81 \\
& Ours-Coarse w/o edges & ours & 1 & 2.08 & 61.39 & 70.42 & 79.05 & 88.59 \\
& Ours-Coarse & ours & 1 & 1.93 & 66.55 & 74.91 & 82.19 & 90.15 \\
& Ours-Fine & ours & 1 & \textbf{1.77} & \textbf{70.27} & \textbf{78.23} & 83.41 & 90.55 \\
\hline

\hline
 & & & & & 2 mm & 4 mm & 7 mm & 10 mm \\
\cline{6-9}
\multirow{13}{*}{Hand}
& McCouat \textit{et al.}~\cite{mccouat2022contour}\# & CVPR-22 & 550 & 0.64 & 96.63 & 99.48 & 99.89 & 99.91 \\
\cline{2-9}
& RPR-Net~\cite{lei2021contrastive}\# & MICCAI-21 & 1 & 5.84 & 28.45 & 54.35 & 75.36 & 86.24 \\
& SAM~\cite{yan2022sam}\# & TMI-22 & 1 & 1.74 & 74.30 & 91.99 & 98.05 & 99.37 \\
& Yin \textit{et al.}~\cite{yin2022one}* Stage I & ECCV-22 & 1 & 2.13 & 60.93 & 89.43 & - & 99.21 \\
& Yin \textit{et al.}~\cite{yin2022one}* Stage II & ECCV-22 & 1 & 1.82 & 66.39 & 92.93 & - & \textbf{99.97} \\
& UOD~\cite{zhu2023uod}* & MICCAI-23 & 1 & 2.52 & 53.37 & 84.27 & - & 97.59 \\
& DDPM~\cite{di_via_2025_wacv}* & WACV-25 & 1 & 28.75 & 27.87 & 46.44 & - & 58.75 \\
& FM-OSD~\cite{miao2024fmosd}* & MICCAI-24 & 1 & 1.41 & \textbf{86.66} & 96.66 & - & 99.11 \\
& KAN-OSD~\cite{tian2026kanosd}* & ISBI-26 & 1 & 1.65 & 79.43 & 94.82 & -  & 98.66 \\
& Ours-Coarse w/o bias & ours & 1 & 1.49 & 81.22 & 96.07 & 99.25 & 99.53 \\
& Ours-Coarse w/o edges & ours & 1 & 1.40 & 79.68 & 97.60 & 99.75 & 99.88 \\
& Ours-Coarse & ours & 1 & 1.22 & 85.12 & \textbf{98.17} & \textbf{99.86} & 99.94 \\
& Ours-Fine & ours & 1 & \textbf{1.18} & 86.52 & 96.77 & 98.84 & 99.24 \\
\hline

\hline
 & & & & & 3 px & 6 px & 9 px & 12 px \\
\cline{6-9}
\multirow{10}{*}{Chest}
& McCouat \textit{et al.}~\cite{mccouat2022contour}\# & CVPR-22 & 195 & 5.15 & 51.67 & 76.00 & 87.00 & 91.67 \\
\cline{2-9}
& RPR-Net~\cite{lei2021contrastive}\# & MICCAI-21 & 1 & 10.89 & 17.36 & 31.55 & 51.68 & 61.12 \\
& SAM~\cite{yan2022sam}\# & TMI-22 & 1 & 8.83 & 21.00 & 46.33 & 61.33 & 71.33 \\
& Yin \textit{et al.}~\cite{yin2022one}* Stage I & ECCV-22 & 1 & 10.16 & 12.33 & 39.00 & 60.33 & - \\
& Yin \textit{et al.}~\cite{yin2022one}* Stage II & ECCV-22 & 1 & 6.89 & 17.33 & 50.33 & 75.33 & - \\
& DDPM~\cite{di_via_2025_wacv}* & WACV-25 & 1 & 14.99 & 19.92 & 46.34 & 64.63 & - \\
& Ours-Coarse w/o bias & ours & 1 & 7.71 & 32.00 & 59.67 & 80.00 & 87.33 \\
& Ours-Coarse w/o edges & ours & 1 & 7.29 & 27.33 & 54.67 & 74.00 & 84.67 \\
& Ours-Coarse & ours & 1 & 6.37 & 31.00 & 60.67 & 83.33 & 89.67 \\
& Ours-Fine & ours & 1 & \textbf{5.46} & \textbf{32.33} & \textbf{69.33} & \textbf{87.00} & \textbf{93.67} \\
\hline

\hline
 & & & & & 2 mm & 4 mm & 6 mm & 8 mm \\
\cline{6-9}
\multirow{9}{*}{Leg}
& McCouat \textit{et al.}~\cite{mccouat2022contour}\# & CVPR-22 & 150 & 2.68 & 61.50 & 85.86 & 91.50 & 94.21 \\
\cline{2-9}
& RPR-Net~\cite{lei2021contrastive}\# & MICCAI-21 & 1 & 12.89 & 2.58 & 12.78 & 26.54 & 36.45 \\
& SAM~\cite{yan2022sam}\# & TMI-22 & 1 & 6.96 & 10.93 & 36.79 & 56.43 & 70.14 \\
& Yin \textit{et al.}~\cite{yin2022one}\# Stage I & ECCV-22 & 1 & 5.34 & 18.95 & 54.67 & 71.96 & 82.39 \\
& Yin \textit{et al.}~\cite{yin2022one}\# Stage II & ECCV-22 & 1 & 4.86 & 21.38 & 63.14 & 75.36 & 84.63 \\
& Ours-Coarse w/o bias & ours & 1 & 5.67 & 26.21 & 59.64 & 77.50 & 86.71 \\
& Ours-Coarse w/o edges & ours & 1 & 4.00 & 25.71 & 62.29 & 82.43 & 90.71 \\
& Ours-Coarse & ours & 1 & 3.14 & 37.93 & 76.43 & 90.36 & 94.64 \\
& Ours-Fine & ours & 1 & \textbf{2.79} & \textbf{50.00} & \textbf{81.64} & \textbf{92.21} & \textbf{94.86} \\
\toprule
\end{tabular}
\end{threeparttable}

\label{Table:Main}
\end{table*}

\vspace{-5mm}

\section{Experiments}

\subsection{Datasets}

\bheading{BMPLE (Leg)} consists of 190 radiographs collected from two collaborating hospitals. The study has been approved by the Hospitals Committee and carried out in accordance with the Declaration of Helsinki. All of the radiographs have been desensitized. The image sizes range from $2396 \times 4950$ to $3200 \times 8500$, and the pixel spacing lies in $0.13mm \sim 0.16mm$. BMPLE is split into training and test subsets with 120 and 70 radiographs, respectively. The radiographs are annotated by one senior orthopedic surgeon. As illustrated in Fig.~\ref{fig:dataset}, 20 landmarks are defined. For each lower limb, 10 landmarks generate 6 axes, which are used to compute 7 biomechanical angles.

\bheading{Cephalometric (Head)} is a widely-used public dataset for cephalometric landmark detection, containing 400 radiographs, and is provided\footnote{Kaggle, Cephalometric X-Ray Landmarks Detection Challenge, \url{https://www.kaggle.com/jiahongqian/cephalometric-landmarks/discussion/133268}.} in IEEE ISBI 2015 Challenge~\cite{wang2016benchmark}. There are 19 anatomical landmarks labeled by 2 expert doctors in each radiograph. The average of the two annotations is used as the ground truth. The image size is $1935 \times 2400$ and the pixel spacing is 0.1mm. The dataset is split into 150 training and 250 testing images according to the official split.

\bheading{Hand X-ray (Hand)} is a public dataset including 909 hand X-ray images. The setting follows~\cite{ref_scn}. The first 609 images are used for training and the remaining 300 images are used for testing. The image size varies within a small range, so all images are resized to $384 \times 384$. Following Payer \textit{et al.}~\cite{payer2016regressing}, who manually labeled 37 landmarks, we calculate the physical distance by assuming that the distance between two wrist endpoints is 50mm.

\bheading{Chest X-ray (Chest)} is a subset of the Kaggle chest X-ray dataset\footnote{\url{https://www.kaggle.com/datasets/nikhilpandey360/chest-xray-masks-and-labels}} consisting of 279 images. It is curated by Zhu \textit{et al.}~\cite{zhu2021you}, who exclude abnormal cases and annotate six landmarks at the lung boundaries. Since no pixel spacing information is provided, we follow~\cite{zhu2021you} and use pixel distance at fixed resolution ($512 \times 512$) for evaluation.

\subsection{Settings}

\bheading{Metrics.}
Following the official challenge~\cite{wang2016benchmark}, we use mean radial error (MRE) to measure the Euclidean distance between prediction and ground truth, and successful detection rate (SDR) under four radii.

\bheading{Implementation details.}
All models are implemented in PyTorch and trained with Adam using a learning rate of 0.001. The embedding dimension is set to 64, the contrastive temperature is $\tau=10$, and the default SGB parameters are $\alpha=0.07$, $\beta=0.8$, and edge factor $\lambda_e=0.3$. Coarse-stage training uses resized full images; fine-stage training uses landmark-centered crops. The final checkpoints are selected by the lowest validation MRE and evaluated by template matching rather than by a pseudo-label-trained detector.

\bheading{Feature extractor.}
The feature extractor uses a ConvNeXtV2-Tiny backbone with a U-Net-style decoder. A SAM-Med2D image feature is fused into the coarse decoder scale, and $1\times1$ projection layers generate multi-scale embeddings for template-to-target matching.

\subsection{Main Results}

Table~\ref{Table:Main} compares \algname{} with fully supervised detectors and recent annotation-efficient methods on four 2D radiological landmark datasets. We report two main variants: \textbf{Ours-Coarse}, which denotes the first-stage global prediction, and \textbf{Ours-Fine}, which denotes the final result after local refinement. The table also includes bias ablations, which are analyzed in the next subsection.

Overall, \algname{} achieves state-of-the-art performance among one-shot/template-based methods on all four datasets. This is notable because our method uses only one annotated template image, while fully supervised methods typically rely on hundreds of labeled images. On Head, Ours-Fine reduces the MRE from 1.93mm to 1.77mm after refinement, outperforming previous one-shot methods such as SAM, Yin \textit{et al.}, UOD, FM-OSD, and KAN-OSD. It also reaches the same MRE as the classical fully supervised method of Lindner \textit{et al.}, showing that the proposed structure-guided matching can substantially narrow the annotation gap.

On Hand, Ours-Coarse already achieves a strong MRE of 1.22mm, and Ours-Fine further improves it to 1.18mm, obtaining the best one-shot MRE. Compared with FM-OSD, Ours-Fine reduces the MRE from 1.41mm to 1.18mm, while maintaining highly competitive SDRs. Although there remains a gap to heavily supervised detectors trained with 550 labeled images, the result demonstrates that dense self-supervised correspondence can provide accurate landmark localization under extreme annotation scarcity.

The advantage of the proposed global-to-local design is more evident on Chest and Leg, where landmarks are often located on repeated boundaries or elongated anatomical structures. On Chest, Ours-Fine reduces the MRE from 6.37px to 5.46px and improves the 6px SDR from 60.67\% to 69.33\%. This result is competitive with fully supervised methods such as GU$^2$-Net and McCouat \textit{et al.}, despite using only one labeled image. On Leg, Ours-Fine reduces the MRE from 3.14mm to 2.79mm and improves the 2mm SDR from 37.93\% to 50.00\%, approaching the fully supervised McCouat \textit{et al.} result and outperforming the fully supervised Chen \textit{et al.} baseline in MRE.

These results suggest that \algname{} is not merely better than previous one-shot methods, but also competitive with several fully supervised detectors on challenging anatomical structures. The global matcher provides reliable coarse localization, while the local matcher further resolves structural ambiguity around repeated edges and bone contours. This validates the effectiveness of using structure-guided self-supervised matching for annotation-efficient medical landmark detection.

\subsection{Ablation}

Table~\ref{Table:Main} reports four variants of our method. Removing the structure-guided bias gives the weakest coarse-stage results. Adding only the distance bias already improves MRE on most datasets, from 1.49mm to 1.40mm on Hand, 7.71px to 7.29px on Chest, and 5.67mm to 4.00mm on Leg, showing that anatomical distance is an effective prior for contrastive matching.

The edge-aware term further improves the coarse model. Compared with Ours-Coarse w/o edges, the full Ours-Coarse reduces MRE from 2.08mm to 1.93mm on Head, 1.40mm to 1.22mm on Hand, 7.29px to 6.37px on Chest, and 4.00mm to 3.14mm on Leg. The gains are especially clear on Chest and Leg, where many landmarks lie on repeated boundaries or elongated bone contours.

Finally, local refinement consistently improves the full coarse prediction. Ours-Fine reduces MRE from 1.93mm to 1.77mm on Head, 1.22mm to 1.18mm on Hand, 6.37px to 5.46px on Chest, and 3.14mm to 2.79mm on Leg. This confirms that the coarse stage provides a reliable search anchor, while the fine stage resolves remaining local ambiguity without training a pseudo-label detector.

\section{Conclusions and Future Work}

In this paper, we presented \algname, a structure-guided coarse-to-fine self-supervised matching framework for one-shot medical landmark detection.
The method learns dense anatomical correspondence from unlabeled images and uses a single annotated template landmark as the query for target localization.
To overcome the structure-agnostic negative treatment in standard contrastive learning, we introduced a structure-guided bias that reweights negative gradients according to relative distance and edge-aware anatomical relevance.
This bias weakly repels nearby structure-relevant candidates while strongly suppressing distant or off-structure negatives, encouraging the learned feature space to preserve local anatomical structures around landmarks.
We further employed separate global and local encoders, where the global stage provides coarse localization and the local stage refines the prediction in a cropped region.
Experiments on four radiological landmark datasets demonstrate that the proposed bias and coarse-to-fine refinement consistently improve one-shot localization performance.
Future work will explore more explicit modeling of inter-landmark spatial relationships and extend the framework toward universal one-shot landmark detection across broader anatomical regions.

\bibliographystyle{IEEEtran}
\bibliography{reference}

\end{document}